\documentclass[preprint,12pt]{elsarticle}

\usepackage{graphicx}

\newtheorem{tm}{Theorem}
\newtheorem{lm}{Lemma}
\newtheorem{cor}{Corollary}
\newtheorem{pr}{Proposition}
\newtheorem{ex}{Example}
\newcommand{\bsquare}{\hbox{\rule{6pt}{6pt}}}
\def\ci{\perp\!\!\!\!\;\!\perp}

\usepackage{amssymb}

\begin{document}

\begin{frontmatter}



\title{Identifiability of an Integer Modular Acyclic Additive Noise Model 
       and its Causal Structure Discovery} 


\author[label1]{Joe Suzuki}
\ead{suzuki@math.sci.osaka-u.ac.jp}
\author[label2]{Takanori Inazumi\fnref{label3}}
\author[label2]{Takashi Washio\corref{cor1}}
\ead{washio@ar.sanken.osaka-u.ac.jp}
\author[label2]{Shohei Shimizu}
\ead{sshimizu@ar.sanken.osaka-u.ac.jp}

\cortext[cor1]{Corresponding author}
\address[label1]{Department of Mathematics, Graduate School of Science, 
Osaka University, 1-1, Machikaneyamacho, Toyonaka, Osaka 560-0043, Japan}
\address[label2]{Department of Reasoning for Intelligence, the Institute 
of Scientific and Industrial Research, Osaka University, 8-1, Mihogaoka, 
Ibaraki, Osaka 567-0047, Japan}
\fntext[label2]{Currently in NTT Communications Corp.}

\begin{abstract}
The notion of causality is used in many situations dealing with uncertainty.
We consider the problem whether causality can be identified given data set generated 
by discrete random variables rather than continuous ones.
In particular, for non-binary
data, thus far it was only known that causality  can be identified except
rare cases. In this paper, we present necessary and sufficient condition
for an integer modular acyclic additive noise (IMAN) of two variables. In
addition, we relate bivariate  and multivariate  causal identifiability
in a more explicit manner, and develop a practical algorithm to find
the order of variables  and their parent sets. We demonstrate its
performance in applications to artificial data and real world body
motion data with comparisons to conventional methods.
\end{abstract}


\end{frontmatter}


\section{Introduction}

We consider the problem of 
inferring causal relation between two random variables $X,Y$ from a finite number of 
samples that have been generated according to the joint distribution (Spirtes et al. 2000).

Solving the problem in a general setting is rather hard, 
and we need some assumptions to find the causal relation: suppose $X,Y$ are related by 
\begin{equation}\label{eq89}
Y=f(X)+e\ ,
\end{equation}
where $f$ is a function from the range of $X$ to that of $Y$, and the noise $e$
 is independent of $X$,
and suppose further that there is no function $g$ from the range of $Y$ to that of $X$ such that
\begin{equation}\label{eq87}
X=g(Y)+h\ ,
\end{equation}
where the noise $h$ is independent  of $Y$.
Then, we can infer that 
 $X$ causes $Y$ but $Y$ does not cause $X$, and say that the causality is identifiable.
On the other hand, if such a function $g$ exists, then we conclude that we cannot infer causality,  and
 say that $X,Y$ are reversible.
This principle (additive noise model) was proposed by Shimizu et. al,
  who demonstrated that causality can be found if  the joint distribution of $X,Y$
 is not Gaussian when $f$ is a linear, i.e., $f(X)=aX$ with some constant $a$ (LiNGAM)~\cite{Shimizu:2006, Shimizu:2009, Shimizu:2011}.

The same principle applies to searching (acyclic) causal relation 
\begin{equation}\label{eq88}
X_i=f_i(X_1,\cdots,X_{i-1})+e_i
\end{equation}
among random variables
 $X_1,\cdots,X_i$, 
 where 
  $e_i$ is independent of $X_1,\cdots,X_{i-1}$, and
 $f_i$ is a linear function of $X_1,\cdots,X_{i-1}$, $i=1,\cdots,d$. 
 The estimated directed acyclic graph (DAG) is 
 found  from a finite number of samples ~\cite{Dodge:2001, Kano:2003, Shimizu:2008}.
The idea \cite{Mooij:2009} is to find $i$ and $f$ such that 
$$e_i=X_i-f(X_1,\cdots,X_{i-1},X_{i+1},\cdots,X_d)$$ is independent of $X_1,\cdots,X_{i-1},X_{i+1},\cdots,X_d$,
 and remove such an $X_i$ (sink variable); 
 starting from $S_d=\{X_1,\cdots,X_d\}$,
 if we repeat the process (removing a sink variable  from $S_{i}$ to obtain $S_{i-1}$, $i=d-1,\cdots,1$), we obtain an order of $X_1,\cdots,X_d$, and can rename the indexes $i=1,\cdots,d$
 of $X_1\cdots,X_d$ so that Eq. (\ref{eq88})  holds for some $f_1,\cdots,f_d$.

The references~\cite{Hoyer:2009, Zhang:2009, Mooij:2009} address using nonlinear functions as $f$ in Eq. (\ref{eq89}).
In another direction, 
\cite{Lacerda:2008} extended Eq. (\ref{eq88}) to the case:
$$X_i=f_i(X_1,\cdots,X_{i-1},X_{i+1},\cdots,X_d)+e_i\ .$$ 
However, those results assumed that the random variables are continuous. 

This paper addresses the case that the random variables take a finite number of values:
suppose each random variable takes a value in the set ${\cal M}:=\{0,1,\cdots,m-1\}$ for an integer $m\geq 2$,
and define arithmetic over ${\cal M}$ as follows: for $x,y\in {\cal M}$,
$x+y$ takes the value $z\in {\cal M}$ if $m$ divides $x+y-z$.
Such a $z \in {\cal M}$ exists and is unique for any $x,y\in {\cal M}$.
For example, if $m=4$, then
$3+2=1$ in ${\cal M}=\{0,1,2,3\}$. 
When $m=2$,  this amounts to binary data with the exclusive-or arithmetic.
Such random cyclic values are abundant in our daily life. 
For example,  directions in $[N,E,S,W]$, months in $[1,2,\dots,12]$.
For two random variables $X,Y$ that take values in $\cal M$, we consider the additive noise model expressed by Eqs. (\ref{eq89})(\ref{eq87}).
The idea can be extended to the multivariate case using  Eq. (\ref{eq88})
({\it integer modulus acyclic additive noise} (IMAN) model).
Recently, several papers deal with such discrete cases, and we discuss those related results in the next section.

Our contributions in this paper are
\begin{enumerate}
\item to express necessary and sufficient conditions on 
causal identifiability 
in a bivariate IMAN model in terms of the probabilities of $X$ and $e$, and
\item to develop a practical algorithm for identifying a causal structure in a multivariate IMAN model under the identifiability. 
\end{enumerate}
This IMAN often appears in circular/directional statistics~\cite{Fisher:1995, Mardia:2009}. 
This is used in time series analysis of phase angles in the frequency domain. 
It has been extensively used for angular data 
representing an object's shape and motion as observed in ubiquitous sensing systems~\cite{Pons:2010}.

In Section 2, we discuss existing results related to this paper. 
In Section 3, we state  theorems on necessary and 
sufficient conditions of reversibility that is equivalent to
non-identifiability for 
 a bi-variate IMAN, and show examples illustrating those theorems. 
These results show that the causal identifiability 
of an IMAN actually holds except in rare situations. 
In Section 4,
 we propose an algorithm for identifying a causal structure in a multivariate IMAN. 
In Sections 5 and 6, 
we show numerical experiments by using  artificial examples and 
real-world data of human body motions to compare with 
conventional approaches, which suggests that the proposed algorithm is actually useful in many situations.

\section{Related Works and Discussion}

Peters  et. al \cite{Peters:2011a} first considered the IMAN model:
let ${\cal M}:=\{0,1,\cdots,m-1\}$ and ${\cal N}:=\{0,1,\cdots,n-1\}$ with $m,n\geq 2$,
and if we assumed $X,Y$ take values in $\cal M$ and $\cal N$, respectively, then
the causal identifiability is defined by non-existence of
$g: {\cal N}\rightarrow {\cal M}$ in Eq. (\ref{eq87}) such that $h$ is independent of $Y$ assuming existence of 
$f: {\cal M}\rightarrow {\cal N}$ in Eq. (\ref{eq89}) such that $e$ is independent of $X$.

The notions of causal identifiability and reversibility are the same even if  $X,Y$ are discrete.
Let ${\rm supp}(X), {\rm supp}(Y), {\rm supp}(e)$ be the sets of elements $x\in {\cal M}, y\in {\cal N}, y-f(x)\in {\cal N}$ such that
$P(X=x)>0, P(Y=y)>0, P(e=y-f(x))>0$, respectively, and denote the number of elements in set $A$ by $|A|$.
They proved  that for reversibility of a bivariate IMAN model, 
the following conditions are necessary (Theorem~4, \cite{Peters:2011a}): 
\begin{itemize}
\item[(1)] 
$|{\rm supp}(Y)|$ divides $|{\rm supp}(X)|\cdot |{\rm supp}(e)|$.
\item[(2)] If 
$|{\rm supp}(X)|=m$ and $|{\rm supp}(Y)|=n$, then  at 
least one additional equality constraint on $P(X=x)$ and $P(e=y-f(x))$ over $x\in {\cal M}$ and $y\in {\cal N}$ is required.
\end{itemize}
assuming that none of $X$, $Y$, $e$ are uniformly distributed and $f$ is not constant. 

Although the above result suggests that it is unlikely that $X,Y$ are reversible in general situations,
no essence has been captured: exactly when  causality is identified for IMAN ?
We would be very pleased if we had a result on necessary and sufficient conditions of reversibility
in terms of $P(X=x)$ and $P(e=y-f(x))$ over $x\in {\cal M}$ and $y\in {\cal N}$, respectively, and would  feel safe because 
we would know exactly when reversibility occurs beforehand.
We know that earthquakes occur very rare even in Japan but would be much happier if we knew exactly when they occur beforehand.


When  $m=n$,  assuming that
 $f: {\cal M}\rightarrow {\cal M}$  is injective, 
we  derive the necessary and sufficient conditions in the next section.
Thus far, the condition was obtained for $m=n=2$:
either $P(X=0)=P(X=1)=1/2$ or $P(e=0)=P(e=1)=1/2$.
The condition we consider in this paper extends the existing result,
and eventually, the proposed algorithm will have more applications.
We notice that the assumption of injectivity 
 can be 
seen in many situations including the circular/directional problems. 
One of the 
most common cases is that $f$ is a composite function 
of a monotonic periodic function of discrete angles 
and a labeling function of the angles. 
This frequently 
appears in angle relations observed in different 
coordinates in mechanical sensing~\cite{Pons:2010}.

On the other hand, in order to establish relation between bivariate and multivariate causal identifiability,
Perters et. al \cite{Peters:2011b} proposed  $({\cal B},{\cal F})$-identifiable
 functional model classes (IFMOCs):
Suppose that each $F_i$ such that $X_i=F_i(X_1,\cdots,X_{i-1},e_i)$ belongs to a
 subset  ${\cal F}$ of $\{{\mathbb R}^m\rightarrow {\mathbb R}|$ for some   $2\leq m\leq d\}$.
Let  ${\cal F}_{|2}:=\{F\in {\cal F}|F: {\mathbb R}^2\rightarrow {\mathbb R}\}$, and ${\cal P}$ the set of the distribution functions.
Let ${\cal B}$ be any set of $(F,F_X,F_e)\in {\cal F}_{|2} \times {\cal P} \times {\cal P}$ such that $Y=F(X,e)$,
 $e$ is independent of $X$, and $Y$ is not independent of $X$, where $F_X, F_e$ are the distribution functions of $X,e$, respectively.
For example, for the original LiNGAM, we may take the $\cal B$ as the set of $(F,F_X,F_e)$ such that
$F(X,e)=aX+e$, and both of $X,e$ should not be Gaussian.
Then, they prove that if the data generated process belongs to any $({\cal B},{\cal F})$-IFMOC,
 we can identify the exact causal graph from data (Theorem~2). 

Our result in this paper does not contradict to the theorem.
Instead, we show relation between bivariate and multivariate causalities in a more specific manner (Propositions 1 and 2),
and propose a method to find a sink based on bivariate causality verification.
More precisely, we obtain a bi-variate IMAN for any pair of variables 
$\{X_i,X_j\} \subset V$ by conditioning all the other variables except $X_i,X_j$. 
In fact, \cite{Peters:2011b} has not addressed any method to 
find a sink variable uniquely from  bi-variate independence relation between $e_i$ and $\{X_j\}_{j\not=i}$ 
as demonstrated in  DirectLiNGAM~\cite{Shimizu:2009, Shimizu:2011}. 

On the other hand, \cite{Sun:2007} proposed a causal ordering method of binary 
variables. However, its identifiability is not insured, 
and its applicability is limited because the computational 
complexity is rather high.

Recently, \cite{Inazumi:2011} showed a necessary 
and sufficient condition on the bi-variate causal 
identifiability of Eq.(\ref{eq89}) for binary variables.
Given a value of $X$, $P(Y)$ coincides with either $P(e=0)$ or $P(e=1)$ irrespective of the value of $P(X)$.
They showed the reverse model Eq.(\ref{eq87}) satisfying the same condition among $X$, 
$Y$ and $h$ exists if and only if $P(e=1)=P(e=0)$ when 
$0<P(e)<1$. They proposed an efficient algorithm to identify 
a unique causal structure in a multivariate binary acyclic 
additive noise model named BExSAM, {\it i.e.}, Eq.(\ref{eq88}) 
modulo $2$, under the identifiability condition. 
However, BExSAM and its algorithm are not suitable for 
generic modular model. Our study indicates that a nontrivial 
condition different from the uniform $P(e)$ is a necessary 
and sufficient condition for reversibility in some generic 
cases, discussed in the next section, and further establishes 
a generic condition where a uniform $P(e)$ is a necessary 
and sufficient condition for reversibility.

\section{Analysis on Bi-variate Identifiability}\label{suzuki}
We show necessary and sufficient conditions 
for the reversibility of a bi-variate IMAN (1)(2), 
where $X,Y,e$ take values in ${\cal M}$. For simplicity, its modulus $m$ is a prime 
or its power, and $f:{\cal M} \rightarrow {\cal M}$ is injective. 
The notation $P(X=i)=p_i$ and 
$P(e=j)=q_j$ is used for brevity, and $0<p_i<1$ is assumed while $0\leq q_i\leq 1$ for $i,j=0,\cdots,m-1$, 
which does not loose any generality since $X$ is not constant in general situations. 
A typical real example arises from human body motion data 
demonstrated in section 6. We first present 
some lemmas on the reversibility for couple moduli 
to help understanding theorems presented later.

\begin{lm}[Reversibility For $m=2$]\label{exlem}
A bi-variate IMAN modulo $2$ with an injective $f$ is 
reversible if and only if one of the following four 
equalities holds: $p_1=1/2$, $q_1=1/2$, $q_1=0$, $q_1=1$.
\end{lm}
Proof. See Appendix \ref{proofexlem}.\hfill $\bsquare$

\begin{lm}[Reversibility For $m=3$]\label{rev3lem}
A bi-variate IMAN modulo $3$ with an injective $f$ is 
reversible if and only if one of the following five 
equalities holds: $p_0=p_1=p_2$, $q_0=q_1=q_2$, $q_0=1$, 
$q_1=1$, $q_2=1$.
\end{lm}
Proof. See Appendix \ref{proofrev3lem}.\hfill $\bsquare$

\begin{lm}[Reversibility For $m=4$]\label{m=4}
A bi-variate IMAN modulo $4$ with an injective $f$ is 
reversible if and only if either one of the following 
ten holds:
$p_0=p_1=p_2=p_3$, $q_0=q_1=q_2=q_3$,
($q_0=q_2=0$, $p_0=p_2$, $p_1=p_3$),
($q_0=q_2=0$, $q_1=q_3$, $P_2$),
($q_1=q_3=0$, $p_0=p_2$, $p_1=p_3$),
($q_1=q_3=0$, $q_0=q_2$, $P_2$),
$q_0=1$, $q_1=1$, $q_2=1$, $q_3=1$,
where $P_2$ expresses the condition 
$(p_1/p_2=p_3/p_0 \ {\rm or}\  p_1/p_0=p_3/p_2)$.
\end{lm}
Proof. See Appendix \ref{proofrev4lem}.\hfill $\bsquare$

These lemmas are now extended to a theorem on a 
necessary and sufficient condition for bi-variate 
causal reversibility of an IMAN covering more 
generic moduli $m$. Before presenting the theorem, 
we need to introduce the notion of ``{\it balanced 
distribution}'' of $\{p_i\}$. We say $p_0,\dots,p_{m-1}$ 
is balanced with respect to $c$ dividing $m$,
 if all the rows in the following matrix are identical for some constants 
$C_0,\dots,C_{c-1}$ and $g: \{0,1,\cdots,c-1\}\rightarrow \{0,c,\cdots,m-c\}$.
\[\left(\begin{array}{cccc}
p_{0+g(0)}/C_0&p_{0+c+g(0)}/C_0&\dots&p_{0+m-c+g(0)}/C_0\\
p_{1+g(1)}/C_1&p_{1+c+g(1)}/C_1&\dots&p_{1+m-c+g(1)}/C_1\\
\cdots&\cdots&\cdots&\cdots\\
p_{c-2+g(c-2)}/C_{c-2}&p_{2c-2+g(c-2)}/C_{c-2}&\dots&p_{m-2+g(c-2)}/C_{c-2}\\
p_{c-1+g(c-1)}/C_{c-1}&p_{2c-1+g(c-1)}/C_{c-1}&\dots&p_{m-1+g(c-1)}/C_{c-1}\\
\end{array}\right).\]
We denote the condition by $P_c$.
For example, suppose $m=4$ as in Lemma 3, $P_2$ says the rows 
should coincide in either 
$$
\left(
\begin{array}{cc}
p_0/(p_0+p_2)&p_2/(p_0+p_2)\\
p_1/(p_1+p_3)&p_3/(p_1+p_3)\\
\end{array}
\right)\ {\rm or}\ 
\left(
\begin{array}{cc}
p_2/(p_0+p_2)&p_0/(p_0+p_2)\\
p_1/(p_1+p_3)&p_3/(p_1+p_3)\\
\end{array}
\right),
$$ 
corresponding to the $g:\{0,1\}\rightarrow \{0,2\}$ such that either $g(0)=g(1)$ or $g(0)\not=g(1)$, which is equivalent to 
either
${p_0}/{p_1}={p_2}/{p_3}$ or ${p_2}/{p_1}={p_0}/{p_3}$, respectively.
On the other hand, we define $c(q_0,\cdots,q_{m-1})$ by the smallest $c\geq 1$ such that $q_j>0 \Longleftrightarrow q_{j+c}>0$.
For example, for $m=8$,
\begin{enumerate}
\item $q_0,\cdots,q_7>0 \Longrightarrow c(q_0,\cdots,q_7)=1$,
\item $q_0,q_2,q_4,q_6=0, q_1,q_3,q_5,q_7>0 \Longrightarrow c(q_0,\cdots,q_7)=2$,
\item $q_0,q_4=0, q_1,q_2,q_3,q_5,q_6,q_7>0 \Longrightarrow c(q_0,\cdots,q_7)=4$,
\item $q_j=1$ for some $j \Longrightarrow c(q_0,\cdots,q_7)=8$,
\item $q_j=0$ for just one $j \Longrightarrow c(q_0,\cdots,q_7)=8$,
\end{enumerate}
and  for $m=4$ as in Lemma 3, if $q_0=q_2=0$ and $q_1=q_3$, then $c(q_0,q_1,q_2,q_3)=2$.

\begin{tm}[Necessary and sufficient condition for reversibility]\label{th1}
Assume that $m$ is a power of a 
prime number. Let $c:=c(q_0,\dots,q_{m-1})$. Then, $X$ 
and $Y$ are reversible in a bi-variate IMAN modulo $m$ 
if and only if $p_j=p_{j+c}=\dots=p_{j+m-c}$ for all 
$j=0,1,\dots,c-1$ or $(q_j=q_{j+c}=\dots=q_{j+m-c}$ 
for all $j=0,1,\dots,c-1$ and $P_c$).
\end{tm}
Proof. See Appendix \ref{proofth1}.\hfill $\bsquare$\\
Lemmas~\ref{exlem} and \ref{rev3lem} are easily derived 
using this theorem. Furthermore, applying $m=4$ to this 
theorem, we obtain the ten conditions in Lemma~\ref{m=4} 
as follows.
\begin{ex}[$m=4$] \quad\\[-6mm]
\begin{description}
\item[$c=1$:] $p_0=p_1=p_2=p_3$ or $q_0=q_1=q_2=q_3$\\[-6mm]
\item[$c=2$:] 1.\ $q_0=q_2=0$ and 
(($p_0=p_2$ and $p_1=p_3$) or ($q_1=q_3$ and $P_2$))\\
\hspace*{2.3mm}2.\ $q_1=q_3=0$ and (($p_0=p_2$ and $p_1=p_3$) 
or ($q_0=q_2$ and $P_2$))\\[-6mm]
\item[$c=4$:] $q_0=1$ or $q_1=1$ or $q_2=1$ or $q_3=1$\\[-5mm]
\end{description}
\end{ex}

The following two corollaries, which are easily derived from 
Theorem~\ref{th1}, show simpler necessary and sufficient 
conditions under some practical assumptions.
\begin{cor}\label{cr1}
Given a prime number $m$, $X$ and $Y$ in bivariate IMAN 
modulo $m$ are reversible if and only if either of the 
following two conditions are met:\\[-6.5mm]
\begin{description}
\item[$c=1$:] $p_0=\dots=p_{m-1}$ or $q_0=\dots =q_{m-1}$.\\[-6.5mm]
\item[$c=m$:]  $q_k=1$ for some $k$.
\end{description}
\end{cor}
Proof. When $m$ is a prime, $c$ is either 1 or $m$ in 
Theorem~\ref{th1}. If $c=1$, then $P_1$ does not require 
any condition, so that Theorem~\ref{th1} reads either 
$p_0 =\dots =p_{m-1}$ or $q_0 =\dots =q_{m-1}$. If $c=m$, 
which is equivalent to $q_k=1$ for some $k$, 
no condition is required other than this. \hfill $\bsquare$
\begin{cor}\label{cr2}
Given a power of some prime number $m$ and 
$q_0,\dots,q_{m-1}>0$, $X$ and $Y$ in bivariate IMAN 
modulo $m$ are reversible if and only if either 
$(p_0=\dots=p_{m-1}$ or $q_0=\dots =q_{m-1})$.
\end{cor}
Proof. When $q_0,\dots,q_{m-1}>0$, which means $c=1$ by 
the definition of $c(q_0,\dots,q_{m-1})$ and Theorem~\ref{th1}, 
$P_1$ does not require any condition. Thus, Theorem~\ref{th1} 
reads either $p_0 =\dots =p_{m-1}$ or 
$q_0 =\dots =q_{m-1}$.\hfill $\bsquare$\\

These results ensure that the causal identifiability 
of a bi-variate IMAN holds except for a finite number 
of special conditions to occur in practice.
If the modulus $m$ does not meet the condition in 
Theorem~\ref{th1}, there are some cases where the 
reversibility holds even if both $\{p_i\}$ and $\{q_j\}$ are 
nonuniform and nonzero. 
\begin{ex}[$m=6$]\quad \label{ex6}
$r\geq 0$, $q_0=q_2=q_3=q_5, q_1=q_4=rq_0$, 
$p_0=p_2=p_4, p_1=p_3=p_5=rp_0$.
\begin{eqnarray*}
R&=&
\left(\begin{array}{cccccc}
\frac{p_0q_0}{C_0}&\frac{p_1q_5}{C_0}&\frac{p_2q_4}{C_0}&\frac{p_3q_3}{C_0}&\frac{p_4q_2}{C_0}&\frac{p_5q_1}{C_0}\\
\frac{p_0q_1}{C_0}&\frac{p_1q_0}{C_0}&\frac{p_2q_5}{C_0}&\frac{p_3q_4}{C_0}&\frac{p_4q_3}{C_0}&\frac{p_5q_2}{C_0}\\
\frac{p_0q_2}{C_0}&\frac{p_1q_1}{C_0}&\frac{p_2q_0}{C_0}&\frac{p_3q_5}{C_0}&\frac{p_4q_4}{C_0}&\frac{p_5q_3}{C_0}\\
\frac{p_0q_3}{C_0}&\frac{p_1q_2}{C_0}&\frac{p_2q_1}{C_0}&\frac{p_3q_0}{C_0}&\frac{p_4q_5}{C_0}&\frac{p_5q_4}{C_0}\\
\frac{p_0q_4}{C_0}&\frac{p_1q_3}{C_0}&\frac{p_2q_2}{C_0}&\frac{p_3q_1}{C_0}&\frac{p_4q_0}{C_0}&\frac{p_5q_5}{C_0}\\
\frac{p_0q_5}{C_0}&\frac{p_1q_4}{C_0}&\frac{p_2q_3}{C_0}&\frac{p_3q_2}{C_0}&\frac{p_4q_1}{C_0}&\frac{p_5q_0}{C_0}\\
\end{array}\right)\\
&=&
\frac{p_0q_0}{r^2+3r+2}
\left(\begin{array}{cccccc}
1&r&r&r&1&r^2\\
r&r&1&r^2&1&r\\
1&r^2&1&r&r&r\\
1&r&r&r&1&r^2\\
r&r&1&r^2&1&r\\
1&r^2&1&r&r&r
\end{array}\right)
\end{eqnarray*}
We find that $X$ and $Y$ are reversible by 
$g(0)=0, g(1)=4, g(2)=2, g(3)=0, g(4)=4, g(5)=2$.
\end{ex}

In Example 2, reversibility is due to shared parameter $r$, which is consistent with Peters et. al [15] who suggested 
that reversibility requires additional equality condition among $\{p_i\}$ and $\{q_j\}$.

\section{IMAN Algorithm}\label{suzuki}
We assume that there exist $i\in \{1,\cdots,d\}$ and $f$ such that 
$$e_i:=X_i-f(\{X_j\}_{j\not=i})$$
is independent of $\{X_j\}_{j\not=i}$.
In [14], we speculate that the condition (multivariate causal identifiability) reduces to bivariate causal identifiability that
$e_i$ is independent of $X_j$ given $\{X_k\}_{k\not=i,j}$ for all $j\not=i$.
We say that such $X_i$ and a minimal subset of $\{X_j\}_{j\not=i}$ on which $f$ depends are a {\it sink} and a {\it parent set}, respectively.
In this paper, we show in Proposition 1 that the claim is true as long as the probabilities of $e_i$ are positive.
Besides, based on the strong support for bi-variate causal identifiability in Section 3 and Proposition 1, we ignore the reversible cases.
From those considerations, we propose an algorithm to find a unique causal structure in an IMAN from a given modular data set $D$.

\begin{figure}[t]
\begin{center} 
\begin{tabular}{l} 
\hline
input: a modular data set $D$ and $V=\{1,\cdots,d\}$.\\
1. compute a frequency table $FT$ for $D$.\\
2. for $k:=d$ to $1$ do\\
3. \hspace*{2mm} ${i(k)} := {\bf find\_sink}(FT,V)$.\\ 
4. \hspace*{2mm} ${\pi(k)} := {\bf find\_parent}(FT,V,{i(k)})$.\\
5. \hspace*{2mm} remove ${i(k)}$ from $V$,\\
   \hspace*{7mm} and marginalize $FT$ with ${i(k)}$.\\
6. end\\
output: a list $((i(1),\pi(1)),\cdots,(i(d),\pi(d)))$.\\
\hline
\end{tabular} 
\end{center} 
\caption{Main Algorithm}\label{mainalg}
\end{figure} 

Figure 1 outlines the proposed algorithm which is an extension of [17] to cover the IMAN modulo $m\geq 2$.
 The algorithm uniquely find a sink variable, which is different from [9][15].
The first step calculates a frequency table $FT$ from $D$. If we have samples of sufficiently large size,
the values of relative frequency converge to the true probabilities.
Steps 3 and 4 find a sink $i(k)$ and its parent set $\pi(k)\subseteq V\backslash \{i(k)\}$
given $FT$ and $V$, respectively, where $V$ is a subset of $\{1,\cdots,d\}$.
Step 5 removes the estimated sink $i(k)$ from $V$, and update $FT$ so that the frequency values can be 
expressed for the updated set $V$ excluding $i$ (marginalization).
This reduces the size of the model by one in the next cycle.
The entire list $\{(i(k),\pi(k))\}_{k=1}^d$ in the output expresses a DAG structure of the IMAN.

\subsection{Finding Sink and Parent Set}

The proposed method is based on the following observation:

\begin{lm}\rm 
\begin{enumerate}
\item $X\ci\{Y,Z\} \Longrightarrow X\ci Y|Z \wedge X\ci Z|Y $
\item If there is no functional relation between $X,Y,Z$, then the converse is also true.
\end{enumerate}
\end{lm}
(Proof: see \cite{pearl} for example.)

\begin{pr}\rm
Suppose $X_1,\cdots,X_d$ have no deterministic relation. 
Then the following conditions are equivalent:
\begin{enumerate}
\item $X_i$ is a sink
\item $e_i$ is independent of $X_j$ given $\{X_h\}_{h\not=i,j}$ for all $j\not=i$
\item $e_i$ is independent of $\{X_h\}_{h\not=i}$
\end{enumerate}
\end{pr}
(Proof: immediate from Lemma 4).

Proposition 1 implies that we can check that $X_i$ is a sink by verifying  
$e_i$ to be independent of $\{X_h\}_{h\not=i}$, and that finding the sink node is as likely as 
bivariate causal identifiability.

In find\_sink, the conditional probability $P(X_i=x_i|\{X_j=x_k\}_{j\not=i})$ is estimated from
FT (we write the value by $\hat{P}(X_i=x_i|\{X_j=x_k\}_{j\not=i})$).
Suppose that $X_i$ is a sink node. 
Then, the probability of $e_i=X_i-f(\{X_j\}_{j\not=i})$ is the same for
$\{X_j=x_j'\}_{j\not=i}$ and $\{X_j=x_j''\}_{j\not=i}$. If we choose 
$x_i'$ and $x_i''$ such that $\hat{P}(X_i=x_i'|\{X_j=x_k'\}_{j\not=i})$ and 
$\hat{P}(X_i=x_i''|\{X_j=x_k''\}_{j\not=i})$ are maximized, respectively.
Then, it is likely that
\begin{equation}\label{eq4}
x_i'-x_i''=f(\{x'_j\}_{j\not=i})-f(\{x''_j\}_{j\not=i})
\end{equation}
if the sample size $n$ is large.
Let $c$ be the value of (\ref{eq4})
Then, 
the distributions of $X_i$ and  $X_i+c$ given $\{X_j=x_j'\}_{j\not=i}$ and $\{X_j=x_j''\}_{j\not=i}$, respectively,
should be the same if the estimation of $c$ is correct.

To this end, we apply the data to the G-test \cite{sokai} which distinguishes whether
$P(\cdot|A)=P(\cdot|B)$ or not for disjoint events $A,B$ from data.
The G-test calculates the G-value:
\begin{eqnarray}
&&2\sum_k{c_n(C_k\cap A)}\log \{\frac{c_n(C_k\cap A)}{c_n(A)}/\frac{c_n(C_k)}{n}\}\nonumber\\
&+&2\sum_k{c_n(C_k\cap B)}\log \{\frac{c_n(C_k\cap B)}{c_n(B)}/\frac{c_n(C_k)}{n}\}\ , \label{eq90}
\end{eqnarray}
where $c_n(\cdot)$ is the frequency of the event, and $\{C_k\}$ are disjoint events covering the whole events ($\cup_k C_k=\Omega$).
The G-test is more correct than the $\chi^2$-test that calculates an approximation of (\ref{eq90}).
In our case, in order to prove Independence, we compare the values $\hat{P}(\cdot|\{X_j=x_j\}_{j\not=i})$ for  all $\{x_j\}_{j\not=i}$ in ${\cal M}^{d-1}$.

Once the sink $i(k)$ is obtained, we find the parent set $\pi(k)\subseteq V\backslash\{i\}$ assuming that the estimated $i(k)$ is correct.
We find the parent based on the following observation:
\begin{pr}\rm 
Suppose $X_1,\cdots,X_d$ have no deterministic relation. 
Then, for any $i\in V:=\{1,\cdots,d\}$ and $\pi\subseteq V-\{i\}$,
$$X_i\ci X_j|\{X_h\}_{h\not=i,j}, j\in V-\pi-\{i\} \Longleftrightarrow X_i\ci \{X_j\}_{j\in V-\pi-\{i\}}|\{X_h\}_{h\in \pi}$$
\end{pr}
(Proof: immediate from Lemma 4).
To this end, for each $j\not=i(k)$,
we compute $\hat{P}(X_{i(k)},X_j|\{X_h=x_h\}_{h\not=i(k),j})$ for all $\{x_h\}_{h\not=i(k),j}\in {\cal M}^{d-2}$
to test if $X_{(i)}$ and $X_j$ are independent  via the G-test given $\{X_h=x_h\}_{h\not=i(k),j}$.
We see that  $j\in \pi(k)$ if and only if 
$X_{i(k)}$ and $X_j$ are not independent given at least one $\{x_h\}_{h\not=i(k),j} \in {\cal M}^{d-2}$.
We test all the tables of 
$\{x_h\}_{h\not=i(k),j} \in {\cal M}^{d-2}$ 
by multiple comparison tests~\cite{Ben:1995}. 
Throughout this paper, find\_parent uses the significance 
level $\alpha=0.05$ and repeats the procedure for all 
$j\not={i(k)}$ to enumerate all the values  of $\pi(k)$.

\subsection{Computational Complexity}\label{complex}
The table FT is of 
size $m^d$. Hence, the space complexity is $O(m^d)$. 
The critical task in find\_sink is to compute 
$m \times m^{d-1}$ conditional probability tables for 
${i(k)}$, and this is repeated at most $d$ times. 
The critical task in find\_parent is the $m^{d-2}$ 
times computation of $m \times m$ conditional probability 
tables for $j\not=i(k)$, and  this is repeated $d-1$ times at 
most. These are further repeated $d$ times in the main 
algorithm shown in Fig.\ref{mainalg}. Accordingly, the 
total time complexity is $O(d^2m^d)$. We require that the 
data size $n$ is near the size of $FT$, {\it i.e.}, $m^d$. 
Therefore, when $n \simeq m^d$, the complexity is virtually 
$O(d^2n)$ which is comparable or better than previous work. 
For example, DirectLiNGAM~\cite{Shimizu:2009, Shimizu:2011} 
which is one of the most efficient causal inference 
algorithm requires $O(d^3n)$.

\section{Experimental Evaluation}

In this section,
we evaluate the basic performance of our algorithm by using artificial data. 
Let $d$ be the number of variables $\{X_i\}_{i=1}^d$,
$m$ the size of the modulus domain ${\cal M}$,
$n$ the number of samples,
$p_a$ the probability that $X_{i}$ is a parent of 
$X_{j}$ for each $i\not=j$,
and $\{q_i\}_{i=0}^{m-1}$ with $\sum_{i=0}^{m-1}q_i=1$ the 
noise distribution.
For each sink $i$ with parent set $\pi$, $f_{i}$ is a function of $\{X_j\}_{j\in \pi}$.
We add noise $e_i$ to $f_{i}(\{x_j\}_{j\in \pi})$ to obtain $X_{i}=x_{i}$ given $\{X_j=x_j\}_{j\in \pi}$.
By repeating the process for $i=1,\cdots,d$, we obtain one realization $\{X_i=x_i\}_{i=1}^d$.
For simplicity, we assume that all the noise $\{e_i\}_{i=1}^d$ share the  same distribution.
Furthermore, by generating $\{X_i=x_i\}_{i=1}^d$ $n$ times
and randomly changing the indexes $(i)_{1\leq i\leq d}$ of $\{X_i\}_{i=1}^d$ into some $(i(k))_{1\leq k\leq d}$, we obtain
data set $D$.
In our experiments, we estimate $\{i(k),\pi(k)\}_{k=1}^d$ from the data set $D$ obtained above.

\begin{table}[t] 
\caption{Performance under various $d$ and $m$.}
\label{result1} 
\begin{center} 
\begin{tabular}{rlllll} 
\multicolumn{1}{c}{\bf $d \backslash m$}&\multicolumn{1}{c}{2}&\multicolumn{1}{c}{3}&\multicolumn{1}{c}{4}&\multicolumn{1}{c}{5}&\multicolumn{1}{c}{6}\\ 
\hline 
&0.033&0.011&0.006&0.004&0.000\\
\cline{2-6} 2&0.958&0.973&0.985&0.985&1.000\\
\cline{2-6} &0.741&0.740&0.766&0.803&0.960\\
\hline 
&0.040&0.016&0.014&0.034&0.005\\ 
\cline{2-6} 4&0.941&0.957&0.958&0.933&0.967\\
\cline{2-6} &2.66&2.97&3.65&5.24&116.4\\
\hline 
&0.036&0.031&0.211&0.317&0.093\\ 
\cline{2-6} 6&0.927&0.921&0.785&0.747&0.844\\
\cline{2-6} &5.57&10.3&8.3&128.0&8736.\\
\hline 
&0.050&0.228&0.338&0.351&0.247\\ 
\cline{2-6} 8&0.898&0.766&0.744&0.748&0.793\\
\cline{2-6} &11.2&86.0&722.&4476.&552060.\\
\hline 
\multicolumn{6}{l}{{\small Top:$Ero$, Middle:$Acc$ and Bottom:$CT$ in a cell.}}\\
\multicolumn{6}{l}{$q_i$ are uniformly random.}
\end{tabular} 
\end{center} 
\end{table} 
\begin{table}[t] 
\caption{Performance under various $d$ and $m$.}
\label{result3} 
\begin{center} 
\begin{tabular}{rlllll} 
\multicolumn{1}{c}{\bf $d \backslash m$}&\multicolumn{1}{c}{2}&\multicolumn{1}{c}{3}&\multicolumn{1}{c}{4}&\multicolumn{1}{c}{5}&\multicolumn{1}{c}{6}\\ 
\hline 
\cline{2-6} 2&0.014&0.011&0.006&0.008&0.160\\
\cline{2-6} &0.976&0.989&0.989&0.991&0.840\\
\hline 
\cline{2-6} 4&0.020&0.031&0.013&0.012&0.273\\
\cline{2-6} &0.959&0.957&0.947&0.946&0.763\\
\hline 
\cline{2-6} 6&0.025&0.058&0.028&0.033&0.261\\
\cline{2-6} &0.944&0.909&0.901&0.895&0.781\\
\hline 
\cline{2-6} 8&0.044&0.148&0.079&0.092&0.290\\
\cline{2-6} &0.913&0.788&0.825&0.812&0.765\\
\hline 
\multicolumn{6}{l}{{\small Top:$Ero$ and Bottom:$Acc$ in a cell.}}\\
\multicolumn{6}{l}{Two adjacent $q_i$ and $q_{i+1}$ are respectively $p$ and $1-p$.}
\end{tabular} 
\end{center} 
\end{table}

We evaluate performance of IMANN in terms of several measures.
Suppose we estimate $\{i(k),\pi(k)\}_{k=1}^d$ to obtain $\{\hat{i}(k),\hat{\pi}(k)\}_{k=1}^d$
 from D. 
Then, we can obtain the adjacency matrices $B$ and $\hat{B}$ for 
 $\{i(k),\pi(k)\}_{k=1}^d$ and 
$\{\hat{i}(k),\hat{\pi}(k)\}_{k=1}^d$, respectively.
If we change the orders of rows and columns so that the matrix $B$
becomes lower triangular, the matrix $\hat{B}$ does not become lower triangular
unless the estimation is correct. 
Let $UT(B,\hat{B})$ be the number of nonzero elements in 
the upper triangular in $\hat{B}$. 
Then, we define $Ero$ by
\[Ero=\frac{UT(B,\hat{B})}{d(d-1)/2}\ ,\]
which expresses the ratio of the number of inconsistent edges
to the number of the whole possible edges.
This measure has been used in many studies including 
LiNGAM~\cite{Shimizu:2006}. 
We also evaluate in how many elements $B$ and $\hat{B}$ coincide, i.e.,
the ratio of the number of elements matched between  $B$ and $\hat{B}$ to 
the number of the whole elements except the diagonal elements, $d(d-1)$.
This measure ($Acc$) expresses accuracy of the estimated causal structure. 
(The less Ero and the larger ACC, the better performance.)
We also evaluate computation time $CT$ (msec) which expresses its algorithm scalability. 
We randomly executed 1000 trials and took average among them. 
For the experiments, we installed MATLAB R2011a on a Windows 7 machine with 
Xeon W3565 (3.2GHz, 4 core, 8MB cache), 6GB RAM and 500GB HDD.			

Based on the result in section~\ref{suzuki}, we are 
particularly interested in the effects of $m$ and 
$\{q_i\}_{i=0}^{m-1}$ on the 
estimation accuracy. 
The parameter $d$ is also an important factor for scalability. 
Table~\ref{result1} shows the results 
under which each $0<q_i<1$ is set uniformly random, $n=1000$, $p_a=0.5$.
In this case, $\{q_i\}_{i=0}^{m-1}$ can be mutually close by chance 
when $m$, is small. 
According to our Theorem~\ref{th1} and Corollary~\ref{cr2}, 
such a condition violates the bi-variate causal 
identifiability of the IMAN. 
However, the chance is 
reduced as $m$ grows. This is reflected in $Ero$ 
and $Acc$ when $m=5$ and $6$.
Although $m=6$ is not 
a power of a prime number, its causal identifiability 
holds similarly to the other numbers. 
This is consistent with the observation in Example~\ref{ex6} 
and \cite{Peters:2011a}.

On the other hand, if $m$ is large, the size of frequency table $FT$
 relative to $n$ is large, and we might not have enough samples to estimate
 the conditional probabilities from $D$ via the G-test.
Because the FT size is $O(m^d)$, the 
critical size $n$ of $D$ should be at least $m^d$ to compute statistically 
accurate $FT$.
For example, the errors can be seen to be significantly 
large when $(d,m)=(6,4),(8,3)$ ($m^d>n=1000$). 
On the other hand, from Table 1, our algorithm seems to provide practical accuracy  if  $n$ is larger than $m^d$. Also,
we find that $CT$ reflects $O(d^2m^d)$ of the algorithm as analyzed in 
Section~\ref{complex}. 
Table~\ref{result3} indicates 
the results when $q_i=p$, $q_{i+1}=1-p$, $q_j=0$, $j\not=i,i+1$ for $0<p<1$, $n=1000$, and $p_a=0.5$.  
In this case, $\{p_i\}_{i=0}^{m-1}$ are always 
far from any reversible conditions shown in our theorem 
and corollaries.  
Thus,  it is reasonable to think that accuracy is obtained as long as $n$ is large enough compared with $m^d$.


Table~\ref{result4} 
and \ref{result5} show the performance in terms of  $n$ and $p_a$ under $d=4$, $m=4$, and 
uniform $q_i$. 
The error is reduced as $n$ grows. However, again we observe 
the critical size of $n$ is  $n \simeq m^d$. 
When $p_a$ is large (the structure is dense), wrong selection of 
of a sink variable in causal ordering affects the 
ordering of all the remaining variables. 
Therefore, $Ero$ increases as density grows
whereas $Acc$ does not decrease.

\begin{table}[t] 
\caption{Performance under various $n$.}
\label{result4} 
\begin{center} 
\begin{tabular}{rlllll} 
\multicolumn{1}{c}{\bf $n$}&\multicolumn{1}{c}{100}&\multicolumn{1}{c}{500}&\multicolumn{1}{c}{1000}&\multicolumn{1}{c}{5000}&\multicolumn{1}{c}{10000}\\ 
\hline 
\cline{1-6} Ero&0.218&0.037&0.014&0.003&0.001\\
\cline{1-6} Acc&0.769&0.923&0.958&0.982&0.983\\
\hline 
\multicolumn{6}{l}{$q_i$ is uniformly random, $d$=4 and $m=4$.}
\end{tabular} 
\end{center} 
\end{table}
\begin{table}[t] 
\caption{Performance under various $p_a$.}
\label{result5} 
\begin{center} 
\begin{tabular}{rllllll} 
\multicolumn{1}{c}{\bf $p_a$}&\multicolumn{1}{c}{0.0}&\multicolumn{1}{c}{0.2}&\multicolumn{1}{c}{0.4}&\multicolumn{1}{c}{0.6}&\multicolumn{1}{c}{0.8}&\multicolumn{1}{c}{1.0}\\ 
\hline 
\cline{1-7} Ero&0.000&0.005&0.012&0.016&0.020&0.023\\
\cline{1-7} Acc&0.964&0.961&0.963&0.958&0.957&0.952\\
\hline
\multicolumn{7}{l}{$q_i$ is uniformly random, $d=4$ and $m=4$.}
\end{tabular} 
\end{center} 
\end{table}

\section{Real-world Applications}\label{realapp}
We analyzed human body orientation data of 
MPI08\_Database\footnote{The data is accessible 
at $http://www.tnt.uni-hannover.de/project/MPI08\_$ 
$Database/$.}~\cite{Pons:2010}. Various indoor 
motions of a human measured with eight movie cameras 
and five orientation sensors are stored in the data. 
Five angle sensors are attached to various parts of 
the body between knees and ankles, between wrists 
and hands and between chest and neck, and angles 
measured with respect to a global inertial coordinate 
frame at 40Hz which is suitable to describe human 
body orientation in his/her view. We obtained angles 
$[roll, pitch, yaw]$ from each snapshot output: $roll$ 
is the rotation angle around the rotation axis; 
$pitch$ is the look up angle between the axis and a 
horizontal plain; and $yaw$ is the horizontal direction 
angle of the axis. $roll$ and $yaw$ take cyclic values 
in $[0,2\pi)$, whereas $pitch$ takes values in 
$[-\pi/2,+\pi/2]$. 
\begin{figure}
\begin{center}
\includegraphics[scale=0.45,clip]{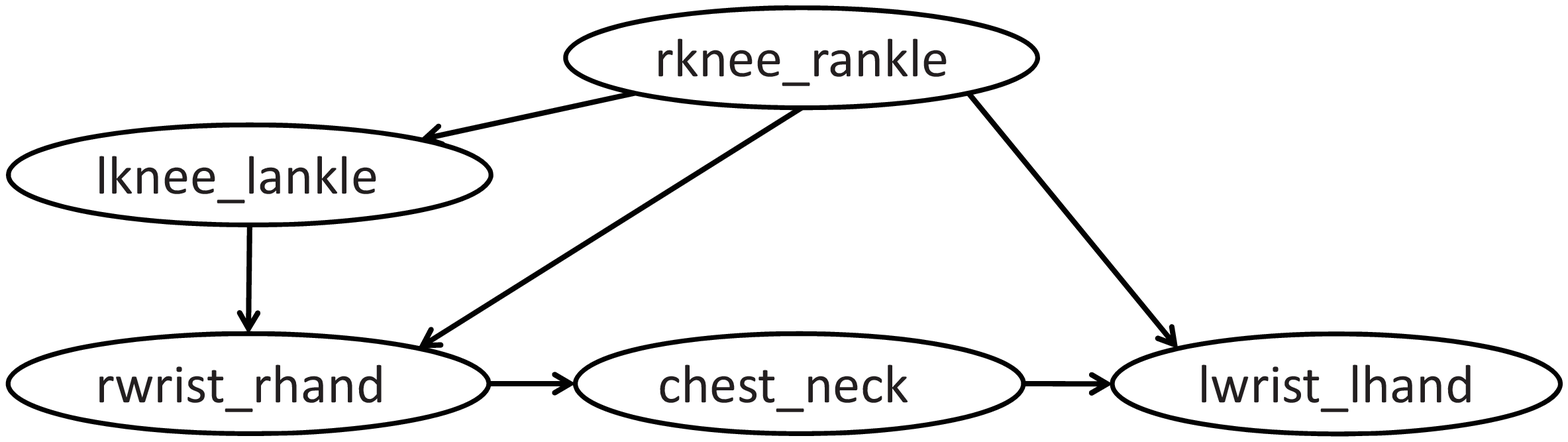}
\caption{A Causal Network (IMAN, $roll$).}\label{eps2}
\end{center}
\begin{center}
\includegraphics[scale=0.45,clip]{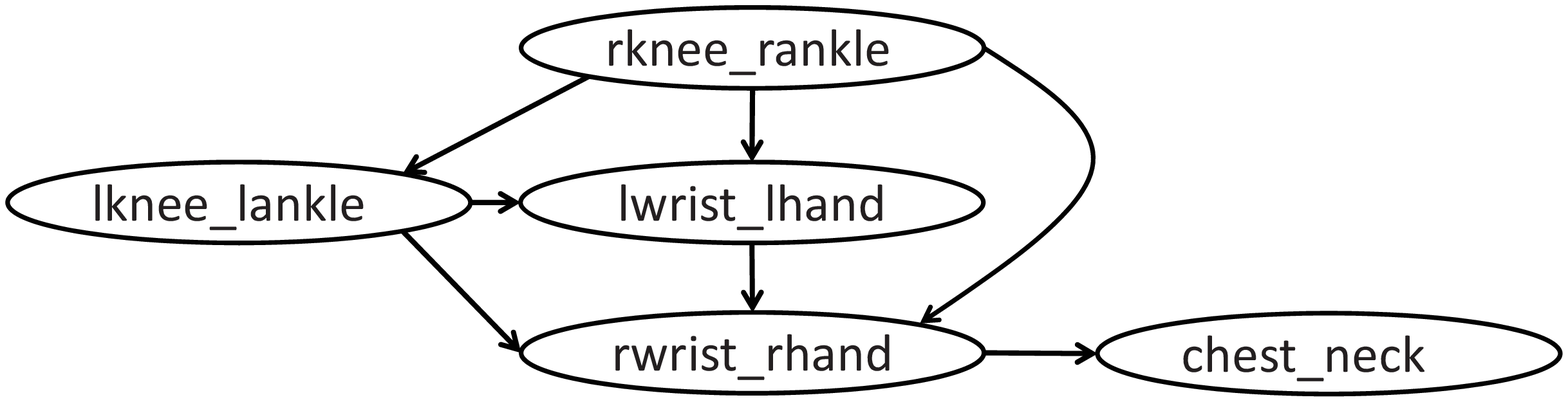}
\caption{A Causal Network (IMAN, $roll$, Cartwheel).}\label{eps3}
\end{center}
\end{figure}

We applied our IMAN algorithm to  $roll$ of the 
data sets; ``ab\_01\_01'' and ``ab\_10\_01.'' 
The former records a counterclockwise walking 
motion over 406 time steps. The latter records two 
cartwheel motions over 580 time steps, and we 
analyzed the first 200 time steps on the leftward 
cartwheel motion. We discretized each $roll$ into 
3 equi-width intervals of $[-\pi,-\pi/6)$, 
$[-\pi/6,+\pi/6)$, $[+\pi/6,+\pi)$. The critical 
$m^d=3^5=243$ is comparable with $n=406$ and $200$ 
for both data sets. Figure~\ref{eps2} shows a 
causal network on ab\_01\_01 by the IMAN algorithm. 
This result is well interpreted in that the right 
leg, its orientation being measured by rknee\_rankle 
senor, takes the initiative, and the left leg 
and right hand, as measured by lknee\_lankle and 
rwrist\_rhand sensors respectively, follow the 
right leg motion, and are further followed by the 
neck and left hand, as measured by chest\_neck and 
lwrist\_lhand sensors respectively. Figure~\ref{eps3} 
shows the result on ab\_10\_01. To initiate the 
leftward cartwheel, the right leg is used first to 
push off from the floor. The motion then influences 
the orientation of his left leg, and subsequently 
planting of these left hand on the floor hand to 
support the body in the rotation. In turn, the right 
hand is planted in similar manner as the body continues 
to rotate. The neck always follows these motions to 
maintain body balance.

\begin{figure}
\begin{center}
\includegraphics[scale=0.45,clip]{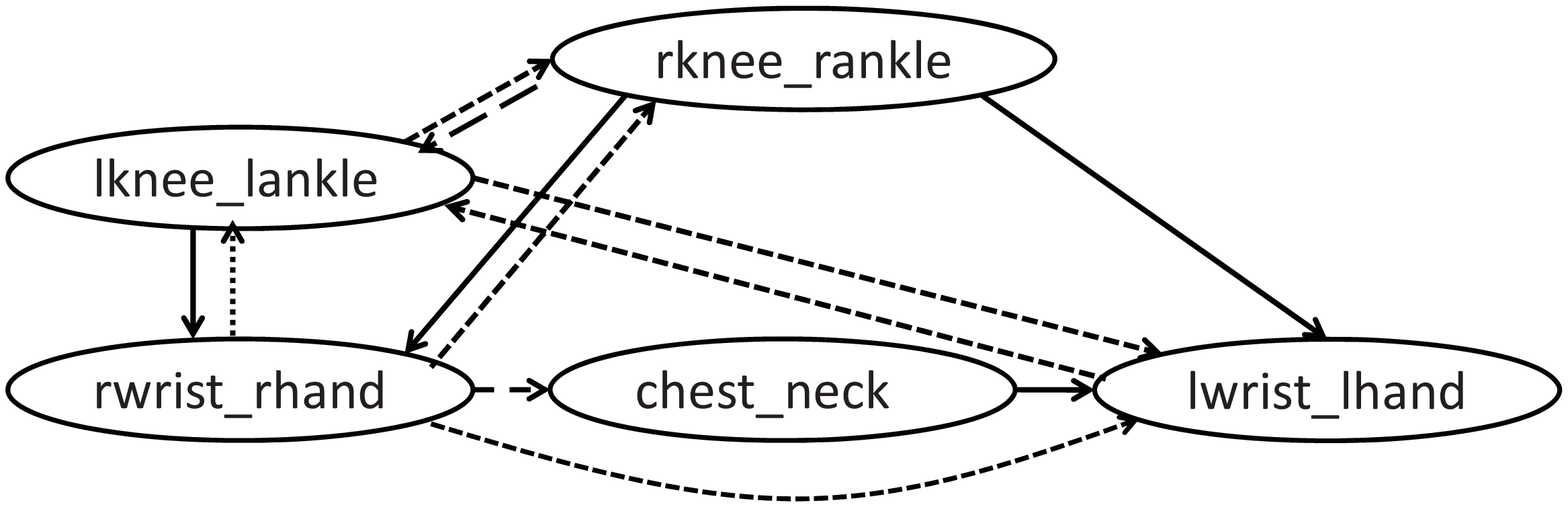}
\caption{A Causal Network (VAR, $pitch$).}\label{eps4}
\end{center}
\begin{center}
\includegraphics[scale=0.45,clip]{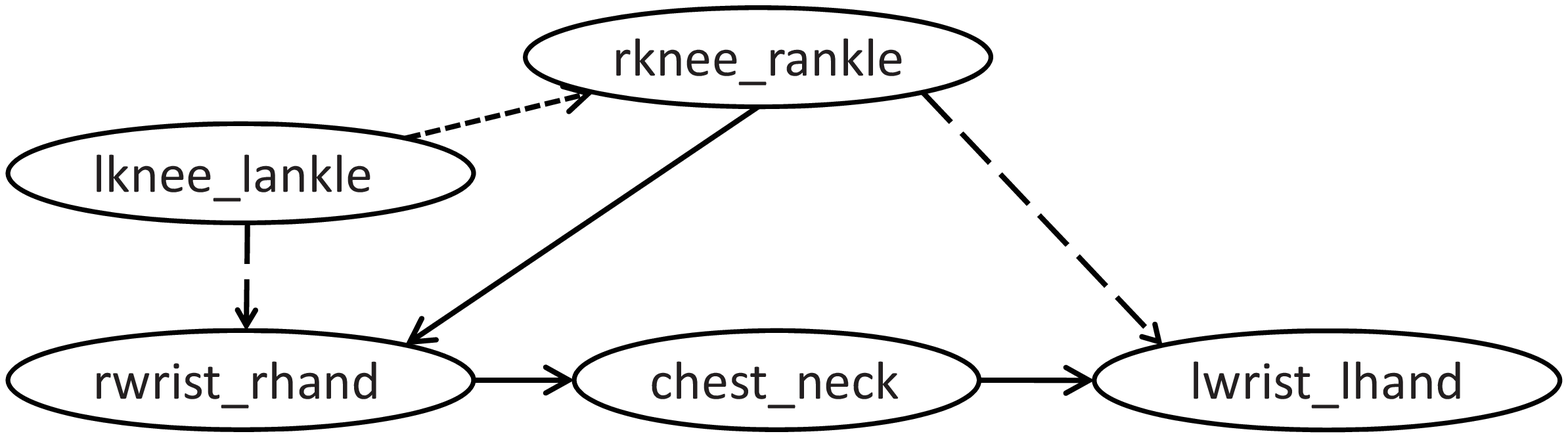}
\caption{A Causal Network (DirectLiNGAM, $pitch$).}\label{eps5}
\end{center}
\end{figure}

Because these are time series representing the body 
motion dynamics,  we also applied VAR (Vector 
Auto-Regressive) (Fig.~\ref{eps4})~\cite{Box:2008} 
and DirectLiNGAM (Fig.~\ref{eps5})~\cite{Shimizu:2009, Shimizu:2011} 
to $pitch$ by assuming $pitch$ shares similar causality 
with $roll$, because no approaches are applicable to 
the modular $roll$. We used the 1st order VAR model 
selected by the final prediction error (FPE) in 
Fig.~\ref{eps4}. In both figures, the causal networks 
are drawn by the matrix elements above a certain threshold 
level. The dotted edges are outputs of VAR and 
DirectLiNGAM that is not in the IMAN output; the dashed 
edges are the output of IMAN but not in the others. 
Though VAR indicates cycles, 4 out of 6 edges of IMAN 
are supported. DirectLiNGAM's causal order is consistent 
except lknee\_lankle. Though these are not from $roll$, 
they are quite consistent with IMAN. We analyzed some 
other data obtained from kicking motions, and produced 
a similar consistency.

\section{Discussion and Conclusion}

In this paper, we presented necessary and sufficient conditions for 
bivariate causal identifiability in IMAN, and actually affirm that 
causality can be identified except in rare cases.
Our result locates exactly when the reversible cases occur.
In addition, we relate bivariate and multivariate causal identifiability in 
more precise manner for IMANN (Propositions 1 and 2).

As a result, we developed a practical way to find a sink and its parent set.
The algorithm needs a sufficient number of samples compared with $m^d$ to verify 
independence when each of $X_1,\cdots,X_d$ takes a value among $m$ values.
The computational complexity is $O(d^2m^d)$, and it is reasonable to say that the value $d$
should be at most 10 for small $m$, which is according to our experiments in this paper
(the value $m$ can be small in practical situations by reducing the quantization level).

If the sample size $n$ is small, we need to improve estimation of FT.
One possibility is to construct a Bayesian measure to deal with each sample set,
and we expect to obtain more robust results even for small $m$.
Then, we can avoid checking independence via the G-test.

The most important direction for future study is 
to seek a causal model and its causal identifiability 
conditions on continuous and cyclic data, such as that 
used in section~\ref{realapp}. If we address these issues, 
our approach can be extended so that we do not need to discretize these data to integers. 
Recent studies in directional statistics provided 
some analyses on distributions of circular/directional 
variables~\cite{Fisher:1995, Mardia:2009}. 

\appendix

\section{Proof of Lemma~\ref{exlem}}\label{proofexlem}
There are four functions $f: {\cal M} \rightarrow {\cal M}$, 
where only $f(X)=X$ and $f(X)=X+1$ are injective.
Let $P(X|Y)$ be such that 
\begin{equation}
P(X|Y):=
\left(
\begin{array}{cc}
P(X=0|Y=0)&P(X=1|Y=0)\\
P(X=0|Y=1)&P(X=1|Y=1)
\end{array}
\right).\label{eq1}
\end{equation}
If $X$ and $Y$ are reversible, there exists $g: {\cal M}\rightarrow {\cal M}$ such that $X=g(Y)+h$, 
$Y \ci h$, which is  equivalent to 
$P(h)=P(h|Y)=P(X-g(Y)|Y)$. Accordingly, 
\[P(h):= \left(
\begin{array}{cc}
P(X-g(0)=0|Y=0)&P(X-g(0)=1|Y=0)\\
P(X-g(1)=0|Y=1)&P(X-g(1)=1|Y=1)
\end{array}
\right),\]
where the first and second rows are mutually identical. 
Note that the upper row is equal to the upper row cyclically 
shifted to the left by $g(0)$ in Eq.(\ref{eq1}) whereas the 
lower row is the $g(1)$ cyclical left shift of the lower 
row in Eq.(\ref{eq1}). If $g(Y)=Y$, this condition is 
equivalent to 
\begin{equation}
P(h):=\left(
\begin{array}{cc}
P(X=0|Y=0)&P(X=1|Y=0)\\
P(X=1|Y=1)&P(X=0|Y=1)
\end{array}
\right).\label{eq2}
\end{equation}
If $g(Y)=Y+1$, $P(h)$ is equal to the matrix in which $X=0$ and 
$X=1$ are exchanged in the above. If $g(Y)=0$, 
\begin{equation}
P(h):=\left(
\begin{array}{cc}
P(X=0|Y=0)&P(X=1|Y=0)\\
P(X=0|Y=1)&P(X=1|Y=1)
\end{array}
\right),\label{eq3}
\end{equation}
and if $g(Y)=1$, $P(h)$ is equal to the matrix in which $X=0$ 
and $X=1$ are exchanged in the above.

To establish the proof, we first consider the case 
$q_1\not=0,1$. For $f(X)=X$, if $X$ and $Y$ are 
reversible under $g(Y)=Y$, the two rows in Eq.(\ref{eq2}) 
are mutually identical:
\begin{eqnarray*}
&&\left(
\frac{(1-p_1)(1-q_1)}{(1-p_1)(1-q_1)+p_1q_1}, \frac{p_1q_1}{(1-p_1)(1-q_1)+p_1q_1}
\right)\\
&&=
\left(
\frac{p_1(1-q_1)}{(1-p_1)q_1+p_1(1-q_1)}, \frac{(1-p_1)q_1}{(1-p_1)q_1+p_1(1-q_1)}
\right)\\
&&\Longleftrightarrow p_1=1/2.
\end{eqnarray*}
Under $g(Y)=Y+1$, $p_1=1/2$ is obtained as well. When $g(Y)=0$, 
the two rows in Eq.(\ref{eq3}) are identical:
\begin{eqnarray*}
&&\left(
\frac{(1-p_1)(1-q_1)}{(1-p_1)(1-q_1)+p_1q_1}, \frac{p_1q_1}{(1-p_1)(1-q_1)+p_1q_1}
\right)\\
&&=
\left(
\frac{(1-p_1)q_1}{(1-p_1)q_1+p_1(1-q_1)}, \frac{p_1(1-q_1)}{(1-p_1)q_1+p_1(1-q_1)}
\right)\\
&&\Longleftrightarrow q_1=1/2.
\end{eqnarray*}
For $g(Y)=1$, $q_1=1/2$ is obtained as well.
For $f(X)=X+1$, the matrices $P(h)$ are the same as that for 
$f(X)=X$ except that the order of the two rows is reversed. 
Hence, the same result is obtained. Thus, as long as $f$ is 
injective and $0<q_1<1$, the reversibility requires either 
$p_1=1/2$ or $q_1=1/2$.

If $q_1=0,1$, meaning $Y=f(X)$ and $Y=f(X)+1$, 
respectively, $X$ and $Y$ are reversible for injective $f$ 
such as $f(X)=X$ and $f(X)=X+1$. \hfill $\bsquare$

\section{Proof of Lemma~\ref{rev3lem}}\label{proofrev3lem}
For injective $f: {\cal M} \rightarrow {\cal M}$,
we wish to find $g: {\cal M} \rightarrow {\cal M}$ 
such that $Y=f(X)+e$ and $h:=X-g(Y)$ are independent.
To this end, $P(h)=P(h|Y)=P(X-g(Y)|Y)$ is obtained by 
cyclic left $g(Y)$ shift of each row in the following 
$P(X|Y)$ to find such $g$.
\[P(X|Y):= \left(
\begin{array}{ccc}
P(X=0|Y=0)&P(X=1|Y=0)&P(X=2|Y=0)\\
P(X=0|Y=1)&P(X=1|Y=1)&P(X=2|Y=1)\\
P(X=0|Y=2)&P(X=1|Y=2)&P(X=2|Y=2)\\
\end{array}
\right).\]
For example, when $g(Y)=Y$,
\begin{eqnarray*}
P(h)&=&P(h|Y)\\
&=&
\left(
\begin{array}{ccc}
P(h=0|Y=0)&P(h=1|Y=0)&P(h=2|Y=0)\\
P(h=0|Y=1)&P(h=1|Y=1)&P(h=2|Y=1)\\
P(h=0|Y=2)&P(h=1|Y=2)&P(h=2|Y=2)\\
\end{array}
\right)\\
&=&P(X-g(Y)|Y)=P(X-Y|Y)\\
&=&
\left(
\begin{array}{ccc}
P(X=0|Y=0)&P(X=1|Y=0)&P(X=2|Y=0)\\
P(X=1|Y=1)&P(X=2|Y=1)&P(X=0|Y=1)\\
P(X=2|Y=2)&P(X=0|Y=2)&P(X=1|Y=2)\\
\end{array}
\right)
\end{eqnarray*}
where the 0th row is a 0 shift of the 0th row of $P(X|Y)$, 
the 1st row is a cyclic left 1 shift of the 1st row of $P(X|Y)$, 
and the 2nd row is a cyclic left 2 shift of the 2nd row of $P(X|Y)$.
However, for ease of notation, we consider the 
following table
\begin{eqnarray*}
P(f(X)|Y)&=&\left(
\begin{array}{c}
P(f(X)=0|Y=0)\\
P(f(X)=0|Y=1)\\
P(f(X)=0|Y=2)\\
\end{array}
\right.\\
&&\hspace*{3mm}\left.
\begin{array}{cc}
P(f(X)=1|Y=0)&P(f(X)=2|Y=0)\\
P(f(X)=1|Y=1)&P(f(X)=2|Y=1)\\
P(f(X)=1|Y=2)&P(f(X)=2|Y=2)\\
\end{array}
\right).
\end{eqnarray*}
Generality is not lost because $f$ must be 
injective. Hereafter, we find $g: {\cal M}\mapsto {\cal M}$ 
such that $f(X)-g(Y)$ and $Y$ are independent. Let 
$$R=P(f(X)|Y)=(r_{k,i})$$ such that 
$$\displaystyle r_{k,i}:=P(f(X)=i|Y=k)=\frac{p_iq_{k-i}}{\sum_{j}p_jq_{k-j}}\ ,$$ where 
$e=Y-f(X)=k-i$ and $f(X) \ci e$.
Thus,
$$R=\left(
\begin{array}{ccc}
p_0q_0/C_0&p_1q_{2}/C_0&p_2q_{1}/C_0\\
p_0q_1/C_1&p_1q_0/C_1&p_2q_{2}/C_1\\
p_0q_2/C_2&p_1q_1/C_2&p_2q_0/C_2
\end{array}
\right)\ ,
$$
where $C_0=p_0q_0+p_1q_{2}+p_2q_{1}$, 
$C_1=p_0q_1+p_1q_{0}+p_2q_{2}$, and 
$C_2=p_0q_2+p_1q_{1}+p_2q_{0}$.

First of all, we consider the case $q_0,q_1,q_2>0$. 
We find conditions that $X$ and $Y$ are reversible for 
each of the nine cases $\{(i,j)\in {\cal M}^2|g(1)=g(0)+i,g(2)=g(0)+j\}$.
If $g(1)=g(0)$, the 0th and the 1st rows in $R$ are 
equally cyclic left-shifted by $g(0)=g(1)$, and thus 
$p_0q_0/C_0=p_0q_1/C_1$, $p_1q_2/C_0=p_1q_0/C_1$, and 
$p_2q_1/C_0=p_2q_2/C_1$. These yield 
$$\frac{p_0q_0}{p_0q_1}=\frac{p_1q_{2}}{p_1q_0}=\frac{p_2q_{1}}{p_2q_{2}} \Longleftrightarrow q_0=q_1=q_2.$$
If $g(1)=g(0)+1$, we cyclically shift 
the 1st row to the left by one column to 
compare with the 0th row. This yields 
$$\frac{p_0q_0}{p_1q_0}=\frac{p_1q_{2}}{p_2q_{2}}=\frac{p_2q_{1}}{p_0q_1} \Longleftrightarrow p_0=p_1=p_2.$$
If $g(1)=g(0)+2$, we cyclically shift 
the 1st row to the left by two columns to 
compare with the 0th row and obtain
$$\frac{p_0q_0}{p_2q_{2}}=\frac{p_1q_{2}}{p_0q_{1}}=\frac{p_2q_{1}}{p_1q_0}=C_0/C_1.$$
By taking a product of these three terms, we obtain $(C_0/C_1)^3=1$. Thus,
\begin{equation}\label{L1}
\frac{p_0q_0}{p_2q_{2}}=\frac{p_1q_{2}}{p_0q_{1}}=\frac{p_2q_{1}}{p_1q_0}=1.
\end{equation}
Similarly, if we compare the 0th and 2nd rows, we obtain
$$\frac{p_0q_0}{p_0q_2}=\frac{p_1q_{2}}{p_1q_1}=\frac{p_2q_{1}}{p_2q_{0}} \Longleftrightarrow q_0=q_1=q_2,$$
\begin{equation}\label{L2}
\frac{p_0q_0}{p_1q_{1}}=\frac{p_1q_{2}}{p_2q_{0}}=\frac{p_2q_{1}}{p_0q_2} =1,
\end{equation}
and
$$\frac{p_0q_0}{p_2q_0}=\frac{p_1q_{2}}{p_0q_2}=\frac{p_2q_{1}}{p_1q_{1}} \Longleftrightarrow p_0=p_1=p_2$$
for $g(2)=g(0)$, $g(2)=g(0)+1$, and $g(2)=g(0)+2$, respectively.
For the eight cases $(i,j)=(0,0),(0,1),(0,2),(1,0),(1,1),(1,2),(2,0)$ and $(2,2)$,
we immediately find
$p_0=p_1=p_2$ or $q_1=q_2=q_3$. Alternatively, for
$(i,j)=(2,1)$, we combine equations (\ref{L1}) and (\ref{L2}). 
By taking products of 
two terms taken from each equation to cancel out $q_i$, we obtain 
$p_1p_2/p_0^2=1 \Longleftrightarrow p_0^3=p_0p_1p_2$ and similarly 
$p_1^3=p_0p_1p_2$, $p_2^3=p_0p_1p_2$ which give $p_0=p_1=p_2$. 
We also obtain $q_0=q_1=q_2$ in the same way. If we substitute 
either $p_0=p_1=p_2$ or $q_0=q_1=q_2$ into the table $R$, we 
can easily find cyclic left-shifts to make all rows identical.
In summary, there exits $g$ for the reversibility under 
$q_1,q_2,q_3>0$, if and only if any of $p_0=p_1=p_2$ or 
$q_1=q_2=q_3$ hold.

On the other hand, suppose just one of $q_0,q_1,q_2$ is zero. For instance, if 
$q_0=0$, $q_0$ should appear in the same column to make the 
rows identical by the cyclic shift $(i,j)=(1,2)$ and 
$$\frac{p_1q_2}{p_2q_2}=\frac{p_2q_1}{p_0q_1}, \frac{p_1q_2}{p_0q_2}=\frac{p_2q_1}{p_1q_1} \Longleftrightarrow p_0=p_1=p_2$$
is obtained. Similarly, $p_0=p_1=p_2$ is required for each of 
($q_0=0, q_1\not=0,q_2\not=0$),
($q_0\not=0, q_1=0,q_2\not=0$), 
($q_0\not=0, q_1\not=0,q_2=0$).

Suppose just two of $q_0,q_1,q_2$ are zero. 
From $q_0+q_1+q_2=1$, one of $q_0=1, q_1=1$ and $q_2=1$ holds. 
Thus, 
$$
R=\left(\begin{array}{cccc}
1&0&0\\
0&1&0\\
0&0&1\\
\end{array}\right),
\left(\begin{array}{cccc}
0&0&1\\
1&0&0\\
0&1&0\\
\end{array}\right), 
\left(\begin{array}{cccc}
0&1&0\\
0&0&1\\
1&0&0\\
\end{array}\right)\ ,$$
hold respectively. Under cyclic left $g(Y)$ shifting by 
$(i,j)=(1,2)$, the three rows in every $R$ become identical. 
Thus, reversibility holds.\hfill $\bsquare$

\section{Proof of Lemma~\ref{m=4}}\label{proofrev4lem}
We denote 
$$R=\left(\begin{array}{cccc}
p_0q_0/C_0&p_1q_3/C_0&p_2q_2/C_0&p_3q_1/C_0\\
p_0q_1/C_1&p_1q_0/C_1&p_2q_3/C_1&p_3q_2/C_1\\
p_0q_2/C_2&p_1q_1/C_2&p_2q_0/C_2&p_3q_3/C_2\\
p_0q_3/C_3&p_1q_2/C_3&p_2q_1/C_3&p_3q_0/C_3\\
\end{array}\right).$$
Suppose that we have a table $T$ such that all rows are identical
by cyclically shifting the $k$-th row left by $g(k)$ columns 
in the table.

First of all, we consider the case {$q_0,q_1,q_2,q_3>0$}.
If the indices $i$ of $p_i$  coincide in more than one row in $T$, 
we have some of the following conditions.
$$
\frac{p_0q_0}{p_0q_1}=\frac{p_1q_3}{p_1q_0}=\frac{p_2q_2}{p_2q_3}=\frac{p_3q_1}{p_3q_2}
\Longleftrightarrow q_0=q_1=q_2=q_3,$$
$$
\frac{p_0q_0}{p_0q_2}=\frac{p_1q_3}{p_1q_1}=\frac{p_2q_2}{p_2q_0}=\frac{p_3q_1}{p_3q_3}\Longleftrightarrow q_0=q_2, q_1=q_3,
$$
$$
\frac{p_0q_0}{p_0q_3}=\frac{p_1q_3}{p_1q_2}=\frac{p_2q_2}{p_2q_1}=\frac{p_3q_1}{p_3q_0}
\Longleftrightarrow q_0=q_1=q_2=q_3,$$
$$
\frac{p_0q_1}{p_0q_2}=\frac{p_1q_0}{p_1q_1}=\frac{p_2q_3}{p_2q_0}=\frac{p_3q_2}{p_3q_3}
\Longleftrightarrow q_0=q_1=q_2=q_3,$$
$$
\frac{p_0q_1}{p_0q_3}=\frac{p_1q_0}{p_1q_2}=\frac{p_2q_3}{p_2q_1}=\frac{p_3q_2}{p_3q_0}
\Longleftrightarrow q_0=q_2, q_1=q_3,$$
$$
\frac{p_0q_2}{p_0q_3}=\frac{p_1q_1}{p_1q_2}=\frac{p_2q_0}{p_2q_1}=\frac{p_3q_3}{p_3q_0}
\Longleftrightarrow q_0=q_1=q_2=q_3.$$
Suppose $q_0=q_2, q_1=q_3$. Then, we have
$$R=
\left(
\begin{array}{cccc}
p_0q_0/C_0&p_1q_1/C_0&p_2q_0/C_0&p_3q_1/C_0\\
p_0q_1/C_1&p_1q_0/C_1&p_2q_1/C_1&p_3q_0/C_1\\
\end{array}\right)$$
written by excluding the identical rows.
If the 1st row cyclically shifted left by $0,1,2,3$ columns is 
identical to the 0th row, we obtain 
$q_0=q_1=q_2=q_3$, $p_0=p_1=p_2=p_3$, ($p_0=p_2, p_1=p_3$),
$p_0=p_1=p_2=p_3$, respectively.  
If $p_0=p_2, p_1=p_3$, then , by excluding the identical columns, 
$$R=\left(
\begin{array}{cc}
p_0q_0/C_0&p_1q_1/C_0\\
p_0q_1/C_1&p_1q_0/C_1\\
\end{array}\right)
\Longleftrightarrow q_0=q_1 \mbox{ or } p_0=p_1$$
upon cyclically shifting left by $0$ or $1$ respectively.
Thus, in any eventuality, $p_0=p_1=p_2=p_3$ or $q_0=q_1=q_2=q_3$ 
ensure the existence of $g$ for reversibility.
From the symmetry of $\{p_i\}$ and $\{q_j\}$, the results where 
$\{p_i\}$ and $\{q_j\}$ are exchanged are obtained even when 
the indices $j$ of $q_j$ coincide in more than one row in $T$. 

Hence, without loss of generality, we compare rows such that 
no $i$ of $p_i$ and $j$ of $q_j$ are the same in any two rows 
in  $T$. For two rows $k,l=0,1,2,3$, there exist columns $i,j$ 
such that $i\not=j$, $k-i\not=l-j$ and 
$$
\frac{p_{i}q_{k-i}}{p_{j}q_{l-j}}=\frac{p_{i+1}q_{k-i-1}}{p_{j+1}q_{l-j-1}}=\frac{p_{i+2}q_{k-i-2}}{p_{j+2}q_{l-j-2}}=\frac{p_{i+3}q_{k-i-3}}{p_{j+3}q_{l-j-3}}
$$
for the identity. If we fix $k$ and multiply the terms 
over rows $l=0,1,2,3$ vertically under the identity of 
all rows for the reversibility, we obtain
\begin{eqnarray*}
&&\hspace*{-7mm}\frac{[p_{i}q_{k-i}]^4}{\prod_up_u\prod_vq_v}=\frac{[p_{i+1}q_{k-i-1}]^4}{\prod_up_u\prod_vq_v}=\frac{[p_{i+2}q_{k-i-2}]^4}{\prod_up_u\prod_vq_v}=\frac{[p_{i+3}q_{k-i-3}]^4}{\prod_up_u\prod_vq_v}
\end{eqnarray*}
where all the denominators are the same since no $i$ of $p_i$ 
and $j$ of $q_j$ are the same in any two rows in T and 
$(i,k-i)$ and $(j,l-j)$ are different for each of $k,l=0,1,2,3$. 
Thus, 
$$p_iq_{k-i}=p_{i+1}q_{k-i-1}=p_{i+2}q_{k-i-2}=p_{i+3}q_{k-i-3}$$
are constant. If we sum over the terms over $k=0,1,\cdots,3$, we obtain
$p_i=p_{i+1}=p_{i+2}=p_{i+3}$,
thus both
$p_0=p_1=p_2=p_3$ and $q_0=q_1=q_2=q_3$ are required for reversibility.


Suppose just one of the $q_0,q_1,q_2,q_3$ are zero. If $q_0=0$, 
$R$ is cyclically shifted to be the following $S$, where all 
the rows are identical for the reversibility.
\begin{eqnarray*}
&&S=\left(\begin{array}{cccc}
0&q_3p_1/C_0&q_2p_2/C_0&q_1p_3/C_0\\
0&q_3p_2/C_1&q_2p_3/C_1&q_1p_0/C_1\\
0&q_3p_3/C_2&q_2p_0/C_2&q_1p_1/C_2\\
0&q_3p_0/C_3&q_2p_1/C_3&q_1p_2/C_3\\
\end{array}\right)\\
&\Longleftrightarrow&
\frac{q_3p_1}{q_3p_2}=\frac{q_2p_2}{q_2p_3}=\frac{q_1p_3}{q_1p_0}\ ,\ 
\frac{q_3p_1}{q_3p_3}=\frac{q_2p_2}{q_2p_0}=\frac{q_1p_3}{q_1p_1}\ ,\ 
\frac{q_3p_1}{q_3p_0}=\frac{q_2p_2}{q_2p_1}=\frac{q_1p_3}{q_1p_2}\\
&\Longleftrightarrow&p_0=p_1=p_2=p_3.
\end{eqnarray*}
Similarly, $p_0=p_1=p_2=p_3$ is obtained for each of $q_1=0, q_2=0, q_3=0$.

Suppose two of $q_0,q_1,q_2,q_3$ are zero. 
If $q_0=q_1=0$, we consider the rows' identity 
in the following $S$.
\begin{eqnarray*}
&&S=\left(\begin{array}{cccc}
0&0&q_3p_1/C_0&q_2p_2/C_0\\
0&0&q_3p_2/C_1&q_2p_3/C_1\\
0&0&q_3p_3/C_2&q_2p_0/C_2\\
0&0&q_3p_0/C_3&q_2p_1/C_3\\
\end{array}\right)\\
&\Longleftrightarrow&
\frac{q_3p_1}{q_3p_2}=\frac{q_2p_2}{q_2p_3}\mbox{ and }
\frac{q_3p_1}{q_3p_3}=\frac{q_2p_2}{q_2p_0}\mbox{ and }
\frac{q_3p_1}{q_3p_0}=\frac{q_2p_2}{q_2p_1}\\
&\Longleftrightarrow&p_0=p_1=p_2=p_3
\end{eqnarray*}
Similarly, $p_0=p_1=p_2=p_3$ is obtained for each of $q_1=q_2=0, q_2=q_3=0, q_3=q_0=0$. For $q_0=q_2=0$, we consider
\begin{eqnarray*}
&&S=\left(\begin{array}{cccc}
0&q_3p_1/C_0&0&q_1p_3/C_0\\
0&q_3p_2/C_1&0&q_1p_0/C_1\\
0&q_3p_3/C_2&0&q_1p_1/C_2\\
0&q_3p_0/C_3&0&q_1p_2/C_3\\
\end{array}\right).
\end{eqnarray*}
The identity of the 0th row and the cyclically 0 or 2 left-shifted 2nd row, 
the identity of the 1th row and the cyclically 0 or 2 left-shifted 3rd row 
and the identity of the 0th row and the cyclically 0 or 2 left-shifted 1st row 
give the following constraints respectively.
\vspace{-1.5mm}
\begin{eqnarray*}
&&\hspace*{-12mm}(\frac{q_3p_1}{q_3p_3}=\frac{q_1p_3}{q_1p_1} \ {\rm or}\  \frac{q_3p_1}{q_1p_1}=\frac{q_1p_3}{q_3p_3})\mbox{ and }(\frac{q_3p_2}{q_3p_0}=\frac{q_1p_0}{q_1p_2} \ {\rm or}\ \frac{q_3p_2}{q_1p_2}=\frac{q_1p_0}{q_3p_0})\mbox{ and}\\
&&\hspace*{-12mm}(\frac{q_3p_1}{q_3p_2}=\frac{q_1p_3}{q_1p_0} \ {\rm or}\  \frac{q_3p_1}{q_1p_0}=\frac{q_1p_3}{q_3p_2})\\
&\Longleftrightarrow&(p_1=p_3\ {\rm or}\  q_1=q_3) \mbox{ and } (p_0=p_2 \ {\rm or}\  q_1=q_3)\\
&& \mbox{ and } (p_1/p_2=p_3/p_0 \ {\rm or}\  q_3p_1/q_1p_0=q_1p_3/q_3p_2)\\
&\Longleftrightarrow&(p_1=p_3 \mbox{ and } p_0=p_2)\ {\rm or}\ (q_1=q_3 \mbox{ and } P_2).
\end{eqnarray*}
Similarly, $q_1=q_3=0 \Longleftrightarrow (p_1=p_3 \mbox{ and } p_0=p_2)\ {\rm or}\ (q_0=q_2 \mbox{ and } P_2)$.

Suppose three of the $q_0,q_1,q_2,q_3$ are zero.
\[R=\left(\begin{array}{cccc}
1&0&0&0\\
0&1&0&0\\
0&0&1&0\\
0&0&0&1\\
\end{array}\right),
\left(\begin{array}{cccc}
0&0&0&1\\
1&0&0&0\\
0&1&0&0\\
0&0&1&0\\
\end{array}\right),
\left(\begin{array}{cccc}
0&0&1&0\\
0&0&0&1\\
1&0&0&0\\
0&1&0&0\\
\end{array}\right),
\left(\begin{array}{cccc}
0&1&0&0\\
0&0&1&0\\
0&0&0&1\\
1&0&0&0\\
\end{array}\right).
\]
Under $g(Y)=i$ cyclically shifting $i$ steps left every 
$i$-th row, the three rows in $R$ become identical. 

The sufficiency of the conditions for all cases 
is easily confirmed by substituting the conditions 
into $R$. \hfill $\bsquare$

\section{Proof of Theorem~\ref{th1}}\label{proofth1}
\noindent (1) When $q_0,\dots,q_{m-1}>0$.\\

Let $R=(r_{k,i})$ be a $m \times m$ matrix such that 
$\displaystyle r_{k,i}=\frac{p_iq_{k-i}}{C_k}$ where 
$C_k=\sum_{i=0}^{m-1}{p_iq_{k-i}} \neq 0$ because 
$p_0,\dots,p_{m-1}>0$ and $\sum_{j=0}^{m-1}q_j=1$. We 
assume that for the reversibility there exist $g(k)$ 
($k=0,\dots,m-1$) so that a shift of every row $k$ to 
the left by $g(k)$ respectively in $R$ derives a matrix 
$T$ where all the rows are identical. 

If $T$ has two rows such that 
$$(p_iq_j/C_{i+j}, p_{i+1}q_{j-1}/C_{i+j}, \dots, p_{i-1}q_{j+1}/C_{i+j})$$
$$(p_iq_k/C_{i+k}, p_{i+1}q_{k-1}/C_{i+k}, \dots, p_{i-1}q_{k+1}/C_{i+k})$$
for some indeces $i,j,k$, then $q_{j+l}/C_{i+j}=q_{k+l}/C_{i+k}$
for $l=0,1,\dots,m-1$. This implies 
$\sum_{l=0}^{m-1} q_{j+l}/C_{i+j} = \sum_{l=0}^{m-1} q_{k+l}/C_{i+k}$. 
Thus, $C_{i+j}=C_{i+k}$ holds because 
$\sum_{l=0}^{m-1} q_{j+l} = \sum_{l=0}^{m-1} q_{k+l} =1$.
Accordingly, $q_{j+l}=q_{k+l}$ holds for $l=0,1,\dots,m-1$, 
which means $q_{k+a}=q_{k}$ for all $k=0,\dots,m-1$ and 
some $a$ dividing $m$. On the other hand, if there exist 
$a$ dividing $m$ such as $q_{k+a}=q_{k}$ for all $k=0,\dots,m-1$, 
then there exist $g(k)$ ($k=0,\dots,m-1$) making $T$ 
 have two rows mentioned above (if $T$ 
does not have such two rows, $q_{k+a}=q_{k}$ for all 
$k=0,\dots,m-1$ hold only for $a:=m$). 

Similarly, if $T$ has two rows such that 
$$(q_jp_i/C_{j+i}, q_{j-1}p_{i+1}/C_{j+i}, \dots, q_{j+1}p_{i-1}/C_{j+i})$$
$$(q_jp_k/C_{j+k}, q_{j-1}p_{k+1}/C_{j+k}, \dots, q_{j+1}p_{k-1}/C_{j+k})$$
for some indeces $i,j,k$, then $p_{i+l}=p_{k+l}$ for 
$l=0,1,\dots,m-1$, which means $p_{k+b}=p_{k}$ 
for all $k=0,\dots,m-1$ and some $b$ dividing $m$ (if $T$ does not have such two rows, $p_{k+b}=p_{k}$ 
for all $k=0,\dots,m-1$ hold only for $b:=m$).

We notice that $\min\{a,b\}$ divides $\max\{a,b\}$ since 
$m$ is a power of some prime number, and that $\max\{a,b\}$ 
divides $m$ by the definitions. Assume the chosen 
$a,b$ are the smallest satisfying the above properties.
Because the values of $\{p_i\}$ and $\{q_j\}$ have cycles of 
$a$ and $b$, respectively, $R$ consists of identical 
$a \times b$ submatrices. Accordingly, we focus on one of 
the submatrices which is sufficient for our proof. 

Let $R=(r_{k,i})$ be such that  $\displaystyle r_{k,i}=\frac{p_iq_{k-i}}{C_k}$ and
 $C_k=\sum_{i=0}^{m-1}{p_iq_{k-i}}$,
and we assume that 
$T$ has been obtained by  shifting row $k$ left by $g(k)$ columns in $R$
so that all the rows are identical in $T$.

If two rows in $T$ are
$$(p_iq_j/C_{i+j}, p_{i+1}q_{j-1}/C_{i+j}, \cdots, p_{i-1}q_{j+1}/C_{i+j})$$
$$(p_iq_k/C_{i+k}, p_{i+1}q_{k-1}/C_{i+k}, \cdots, p_{i-1}q_{k+1}/C_{i+k})$$
for some indeces $i,j,k$, 
then $q_{j+l}=q_{k+l}$ for $l=0,1,\cdots,m-1$,
which means $q_{k+a}=q_{k}$ for some $a$ dividing $m$, else $a:=m$.
On the other hand,  if two rows in  $T$ are
$$(q_ip_j/C_{i+j}, q_{i-1}p_{j+1}/C_{i+j}, \cdots, q_{i+1}p_{j-1}/C_{i+j})$$
$$(q_ip_k/C_{i+k}, q_{i-1}p_{k+1}/C_{i+k}, \cdots, q_{i+1}p_{k-1}/C_{i+k})$$
for some indeces $i,j,k$,
then $p_{j+l}=p_{k+l}$ for $l=0,1,\cdots,m-1$,
 which means $p_{k+b}=p_{k}$ for some $b$ dividing $m$, else $b:=m$.

We notice that $\min\{a,b\}$ divides $\max\{a,b\}$, and  $\max\{a,b\}$ divide $m$,
and assume the chosen $a,b$ are the smallest  satisfying the above properties.

Suppose $a\leq b$.
In matrix 
{\small
$$R=
\left(
\begin{array}{cccc}
p_0q_0/C_0&p_1q_{a-1}/C_0&\cdots&p_{b-1}q_{1}/C_0\\
p_0q_1/C_1&p_1q_0/C_1&\cdots&p_{b-1}q_{2}/C_1\\
\cdots &\cdots&\cdots\\
p_0q_{a-1}/C_{a-1}&p_1q_{a-2}/C_{a-1}&\cdots&p_{b-1}q_{0}/C_{a-1}\\
\end{array}
\right)\ ,
$$
}
if we do not shift row 1  to compare with row 0 in $R$, we obtain
$q_0=\cdots=q_{a-1}$; and 
if we shift row 1 right by $j\not=0$ columns to compare with row 0, we obtain
for $u=0,\cdots,a-1$
$$\frac{p_{j+u}q_{a-j-u}}{p_uq_{1-u}}=\frac{p_{j+u+a}q_{-j-u}}{p_{u+a}q_{1-u-a}}\ .$$
which means for $u=0,1,\cdots,b-1$,
$$\frac{p_{u+a}}{p_u}=\frac{p_{j+u+a}}{p_{j+u}}=\cdots \frac{p_{u-j+a}}{p_{u-j}}=\cdots\ .$$
By multiplying all the terms, we  find that $\displaystyle \frac{p_{u+a}}{p_u}=1$ for $u=0,1,\cdots,b-1$.

If $b\leq a$,  we similarly find that $\displaystyle \frac{q_{u+b}}{q_u}=1$ for $u=0,1,\cdots,a-1$
So, we only need to consider $R$ of size $d\times d$ with $d:=\{a,b\}$, and may assume
that there will be no nontrivial relation among $\{p_i\},\{q_j\}$,
which means that there will be no conflict among indeces.

We complete this proof if we show either
$p_0=\cdots=p_{d-1}$ or $q_0=\cdots = q_{d-1}$, which means 
$p_0=\cdots=p_{m-1}, q_0=\cdots = q_{m-1}$.
However, we see that 
$\displaystyle \prod_{j=0}^{d-1}t_{i,j}=\frac{1}{C_{i+j}^d}\prod_{u=0}^{d-1}p_u\prod_{v=0}^{d-1}q_v$, for $T=(t_{i,j})$.
Since  $\displaystyle \frac{t_{0,j}}{t_{0,k}}=\frac{t_{1,j}}{t_{1,k}}=\cdots=\frac{t_{d-1,j}}{t_{d-1,k}}=C$ and
 $\displaystyle 1=\prod_{i=0}^{d-1}\frac{t_{i,j}}{t_{i,k}}=C^d$, we have
 $\frac{t_{i,j}}{t_{i,k}}=C=1$ for $k=0,\cdots,d-1$.
On the other hand, since
$\displaystyle \prod_{k=0}^{d-1}
\frac{t_{i,j}}{t_{i,k}}
=t_{i,j}^d
/\{\frac{1}{C_{i+j}^d}
\prod_{u=0}^{d-1}p_u\prod_{v=0}^{d-1}q_v=1\}$, we have
$\displaystyle t_{i,j}=\frac{1}{C_{i+j}} {[\prod_{u=0}^{d-1}p_u\prod_{v=0}^{d-1}q_v]^{1/d}}$
for all $i,j=0,\cdots,d-1$. Thus,
$\displaystyle p_iq_j={[\prod_{u=0}^{d-1}p_u\prod_{v=0}^{d-1}q_v]^{1/d}}$
for all $i,j=0,\cdots,d-1$, which means 
$\displaystyle p_i=q_j=d{[\prod_{u=0}^{d-1}p_u\prod_{v=0}^{d-1}q_v]^{1/d}}$
for all $i,j=0,\cdots,d-1$.

In any case, $p_0=\cdots=p_{m-1}$ or $q_0=\cdots = q_{m-1}$ if $q_0,\cdots,q_{m-1}>0$.

\noindent (2) When $q_j=0$ for some $j$.
For each $k$ such that $q_k>0$, we 
select columns $k,k+c,\cdots,k+m-c$ in $S$ to obtain the matrix of size ${m}\times \frac{m}{c}$
$$S_k:=
\left(
\begin{array}{cccc}
\frac{q_kp_{-k}}{C_0}&\frac{q_{k-c}p_{-k+c}}{C_0}&\cdots&\frac{q_{k+c}p_{-k-c}}{C_0}\\
\frac{q_kp_{-k+1}}{C_{1}}&\frac{q_{k-c}p_{-k+1+c}}{C_{1}}&\cdots&\frac{q_{k+c}p_{-k+1-c}}{C_{0}}\\
\cdots&\cdots&\cdots&\cdots\\
\frac{q_kp_{-k+m-1}}{C_{m-1}}&\frac{q_{k-c}p_{-k+m-1+c}}{C_{m-1}}&\cdots&\frac{q_{k+c}p_{-k+m-1-c}}{C_{m-1}}\\
\end{array}
\right)\ .
$$
Furthermore, for each $j=0,1,\cdots,c-1$, we select rows $j,j+c,\cdots,j+m-c$ in $S_k$ to obtain the matrix
of size $\frac{m}{c}\times \frac{m}{c}$
$$S_{jk}:=
\left(
\begin{array}{cccc}
\frac{q_kp_{j-k}}{C_0}&\frac{q_{k-c}p_{-j-k+c}}{C_0}&\cdots&\frac{q_{k+c}p_{j-k-c}}{C_0}\\
\frac{q_kp_{j-k+c}}{C_{1}}&\frac{q_{k-c}p_{j-k+2c}}{C_{1}}&\cdots&\frac{q_{k+c}p_{j-k}}{C_{0}}\\
\cdots&\cdots&\cdots&\cdots\\
\frac{q_kp_{j+m-k-c}}{C_{m-1}}&\frac{q_{k-c}p_{j+m-k}}{C_{m-1}}&\cdots&\frac{q_{k+c}p_{j+m-k-2c}}{C_{m-1}}\\
\end{array}
\right)\ .
$$
Since $S_{jk}$ is a square matrix and all the elements are positive, for reversibility, the condition
$$p_j=p_{j+c}=\cdots=p_{j+m-c}\ {\rm or}\ q_k=q_{k+c}=\cdots=q_{k+m-c}$$
for all $j=0,1,\cdots,c-1$ and $k=0,1,\cdots,c-1$ (for $k$ such that $q_k=0$, the condition is trivially true) is required,
i.e., either
\begin{enumerate}
\item $p_j=p_{j+c}=\cdots=p_{j+m-c}$ for $j=0,1,\cdots,c-1$, or
\item $q_k=q_{k+c}=\cdots=q_{k+m-c}$ for $k=0,1,\cdots,c-1$
\end{enumerate}
is necessary. It remains to prove that if either of the two condition is satisfied, reversibility holds.
Under the first condition, $S_k$ is a uniform matrix, and expresses reversibility.
Under the second condition, in $S_k$ consisting of $\{p_l\}$, the $i$th and $j$th rows coincide each other by shifting if $|i-j|$ is divided by $c$.
Hence, the condition that $S_k$ expresses reversibility  is equivalent to $P_c$ under the second condition.
\hfill $\bsquare$



\bibliographystyle{model1a-num-names}
\bibliography{imannc}



%

\end{document}